\documentclass[runningheads]{llncs}

\usepackage[T1]{fontenc}
\usepackage{graphicx}
\usepackage{amsmath}
\usepackage{amssymb}
\usepackage{booktabs}
\usepackage{multirow}
\usepackage{array}

\newcommand{\method}{RRDA}
\newcommand{\newobj}{\ensuremath{o^\star}}
\newcommand{\oldobj}{\ensuremath{o^{0}}}
\newcommand{\score}{\operatorname{score}}
\newcommand{\softplus}{\operatorname{softplus}}
\newcommand{\cf}{CounterFact}
\newcommand{\zsre}{ZsRE}
\newcommand{\mquake}{MQuAKE-CF}

\begin{document}

\title{When to Write and When to Suppress: Route-Specialized Dual Adapters for Memory-Assisted Knowledge Editing}
\titlerunning{Route-Specialized Dual Adapters}

\author{Baijia Zhang\inst{1} \and
Yining Huang\inst{2} 
}
\authorrunning{Zhang et al.}
%
\institute{Xinjiang University
\email{107552404826@stu.xju.edu.cn}\\
\and
Meta Emergence Laboratory\\
\email{huangyining1987@gmail.com}\\
}

\maketitle

\begin{abstract}
Knowledge editing systems must update selected facts while preserving nearby but irrelevant behavior. This paper studies this problem in a memory-assisted setting where an edit memory is retrieved at inference time and a parameter-efficient adapter corrects the model's object preference. We argue that the central design question is not only how to write an edit, but also when to suppress it. We introduce \method{}, a route-specialized dual-adapter editor. A relevance router first decides whether a prompt should receive an edit memory. Routed prompts use an edit adapter trained to prefer the new object over the original object; unrouted non-direct prompts use a separate locality adapter trained to preserve or restore the original-object preference. We evaluate \method{} on three 1,000-case protocols, \cf{}, \zsre{}, and \mquake{}, under the same memory protocol and two 7B/8B base models. On Llama-3.1-8B-Instruct, \method{} obtains the best overall probability-preference accuracy on all three benchmarks: 0.8180 on \cf{}, 0.8946 on \zsre{}, and 0.9922 on \mquake{}. The same trend holds on Qwen3-8B. Router ablations show that the relevant memory boundary differs across datasets: a lexical neural router is safest on \cf{}, while BGE embedding routing is better on \zsre{} and \mquake{}. Memory, component, and module ablations show that explicit memory supplies the largest edit gain, while route-specialized adapters improve the final reliability-locality balance rather than simply increasing LoRA capacity.
\keywords{Knowledge editing \and Parameter-efficient fine-tuning \and LoRA \and Retrieval \and Locality \and Large language models}
\end{abstract}

\section{Introduction}

Large language models store factual associations that may become outdated, wrong, or inconsistent with a target application. Knowledge editing aims to change such associations after pretraining without retraining the entire model \cite{zhu2020modifying,decao2021editing,mitchell2022mend,meng2022rome}. A useful editor should be reliable on the edited prompt, generalize to paraphrases, and remain local: nearby prompts that mention related subjects or relations should not inherit the edit.

Locality is the hard part. In parametric editing, a local weight update can over-generalize to neighboring facts \cite{meng2023memit,gupta2025normgrowth}. In memory-based editing, a retrieved edit can be injected into prompts where it does not belong \cite{mitchell2022serac,zheng2023ike,wang2024wise}. Parameter-efficient fine-tuning (PEFT) does not remove this problem. A LoRA adapter \cite{hu2022lora} trained to promote the new object can be helpful on relevant prompts but unsafe if it is activated for an irrelevant prompt. Thus a memory-assisted editor must solve two linked decisions: when should the edit memory be retrieved, and which correction module should be active after retrieval?

This paper builds around that observation. We use a probability-preference view of editing. Each case has a subject, a relation, an original object \oldobj{}, and a desired new object \newobj{}. On direct and paraphrase prompts, success means that the model assigns a higher normalized log-probability to \newobj{} than to \oldobj{}. On locality prompts, success means that the protected or original object remains preferred. This metric is stricter than generation no-leak because it catches cases where the edited object has silently become more probable even if greedy decoding does not yet reveal it.

We propose \method{} (Relevance-Routed Dual Adapters), a small memory-assisted editor that separates writing from suppression. A router chooses whether an input is relevant to a stored edit memory. If the prompt is direct or routed as relevant, \method{} inserts the memory and activates an edit LoRA adapter. If the prompt is not routed and is non-direct, \method{} does not insert memory and activates a separate locality adapter. The edit adapter is trained with edited-answer cross-entropy and pairwise ranking. The locality adapter is trained only to restore locality preferences. The route policy is fixed after validation, which makes the inference path deterministic and auditable.

The experiments support a broader route-specialization story. First, router ablations show that no single relevance rule is best everywhere. \cf{} clean prompts favor subject or lexical neural routing; stress tests expose subject-only routing failures; \zsre{} favors semantic embedding routing because locality prompts often share the subject but change the relation; \mquake{} favors semantic retrieval because multi-hop questions differ lexically from stored single-hop edits. Second, the main editor comparison uses three benchmarks, three method families, and two base models. Across \cf{}, \zsre{}, and \mquake{}, \method{} improves overall probability-preference accuracy over resource-matched LocFT-style and PRUNE-style PEFT baselines on both Llama-3.1-8B-Instruct \cite{dubey2024llama} and Qwen3-8B \cite{yang2025qwen3}. Third, ablations show why: memory contribution ablations separate the effect of explicit memory from adapters; component ablations isolate the locality adapter's role; and module sweeps show that expanding LoRA targets beyond q/v gives little benefit.

Our contributions are:
\begin{itemize}
  \item We formulate memory-assisted PEFT editing as a routed probability-preference problem, with one path for writing edits and another for suppressing off-route edit leakage.
  \item We introduce \method{}, a route-specialized dual-adapter editor that uses an edit adapter for routed prompts and a locality adapter for unrouted non-direct prompts.
  \item We provide a 3-benchmark, 3-method, 2-base-model evaluation under a shared memory protocol, plus memory contribution, router, adapter, and target-module ablations.
  \item We clarify the limitations of the current protocol, especially that our \mquake{} evaluation is probability-preference with synthetic locality rather than the official exact-match generation benchmark.
\end{itemize}

\section{Related Work}

\paragraph{Knowledge editing.}
Early work edited factual behavior through constrained fine-tuning or learned update networks \cite{zhu2020modifying,decao2021editing,mitchell2022mend}. ROME localizes factual associations in transformer MLPs \cite{meng2022rome}, and MEMIT extends this idea to many edits \cite{meng2023memit}. Later work studies scaling and stability, including MALMEN \cite{tan2023malmen}, null-space preservation in AlphaEdit \cite{fang2025alphaedit}, perturbation-restrained editing in PRUNE \cite{ma2025prune}, and analyses of norm growth in sequential editing \cite{gupta2025normgrowth}. Our setting is more modest: base weights remain frozen, and edited knowledge is partly supplied by external memory. The question is how to route and correct probabilities safely.

\paragraph{Memory-based and in-context editors.}
SERAC routes inputs to a memory-based counterfactual model \cite{mitchell2022serac}. IKE shows that factual editing can be framed through in-context examples \cite{zheng2023ike}. GRACE, WISE, and MemEIC study memory structures for lifelong or continual editing \cite{hartvigsen2023grace,wang2024wise,seong2025memeic}. These methods highlight an important systems principle: the edit store is useful only when relevance is correctly identified. \method{} follows this hybrid direction, but focuses on route-conditioned low-rank correction rather than a standalone retrieval or reasoning engine.

\paragraph{PEFT-based editing.}
LoRA and QLoRA make low-rank adaptation practical for large models \cite{hu2022lora,dettmers2023qlora}. Recent editing work explores efficient or sparse low-rank updates, including time-sensitive editing \cite{ge2024timesensitive}, RoseLoRA \cite{wang2024roselora}, MEMLA \cite{xie2024memla}, and fine-tuning protocols that make FT competitive for editing \cite{yang2026locft}. In contrast, our adapter is not expected to store all edited facts by itself. The memory carries the edited fact; the adapter changes the probability boundary and the locality adapter suppresses off-route leakage.

\paragraph{Evaluation and routing.}
\cf{} is widely used for factual association editing \cite{meng2022rome}. \zsre{} tests relation extraction style edits \cite{levy2017zeroshot}, and \mquake{} stresses multi-hop consequences after edits \cite{zhong2023mquake}. EasyEdit standardizes reliability, generality, and locality tooling \cite{wang2024easyedit}, while RippleEdits studies logical ripple effects \cite{cohen2024ripple}. Our router experiments also use semantic retrieval tools: BGE embeddings \cite{xiao2023cpack}, sentence-transformer style scoring \cite{reimers2019sentencebert}, and MiniLM cross-encoder reranking \cite{wang2020minilm}. The empirical point is that relevance routing is not universal; different benchmarks reward different no-route boundaries.

\section{Methodology}

\subsection{Problem Setup}

An edit case is represented as $(s,r,\oldobj,\newobj)$: subject, relation, original object, and new object. The memory bank contains one textual memory record per edit case. A query prompt $x$ is direct, paraphrased, or local. Direct and paraphrase prompts should prefer \newobj{}; local prompts should prefer \oldobj{} or a protected locality answer. For a candidate object string $y$, we use length-normalized log-probability:
\begin{equation}
  \score_{\theta,\phi,g}(x,y)
  =
  \frac{1}{|y|}
  \sum_{t=1}^{|y|}
  \log p_{\theta,\phi,g}(y_t \mid x,y_{<t}),
\end{equation}
where $\theta$ is the frozen language model, $\phi$ denotes active LoRA parameters, and $g$ is the adapter gate. Direct and paraphrase prompts are correct when
\begin{equation}
  \score(x,\newobj) > \score(x,\oldobj),
\end{equation}
while locality prompts are correct when the protected object remains preferred. Overall accuracy is the micro-average over all test prompts.

\subsection{Relevance Routing and Memory Construction}

The memory bank is built from direct edit records. \cf{} memory is a direct prompt followed by \newobj{}; \zsre{} memory is a question-answer style record; \mquake{} memory concatenates the requested single-hop rewrites in the composed case. At inference, a router chooses either one memory record or no memory. We evaluate four router families:
\begin{itemize}
  \item \textbf{Subject}: route if the query contains the edited subject string.
  \item \textbf{Subject+relation}: require subject match plus relation-token overlap.
  \item \textbf{NN}: a hashed bag-of-words MLP over query-candidate pairs, trained with positive same-case pairs and random, locality, and hard subject negatives.
  \item \textbf{Semantic}: BGE embedding similarity, optionally followed by MiniLM cross-encoder reranking over the embedding top-$k$.
\end{itemize}
The selected editor router is validation driven: \cf{} uses NN routing; \zsre{} and \mquake{} use BGE embedding routing. The router is not trained to decide which examples enter memory; the memory bank is fixed before evaluation.

\subsection{Route-Specialized Dual Adapters}

\begin{figure}[t]
\centering
\includegraphics[width=\textwidth]{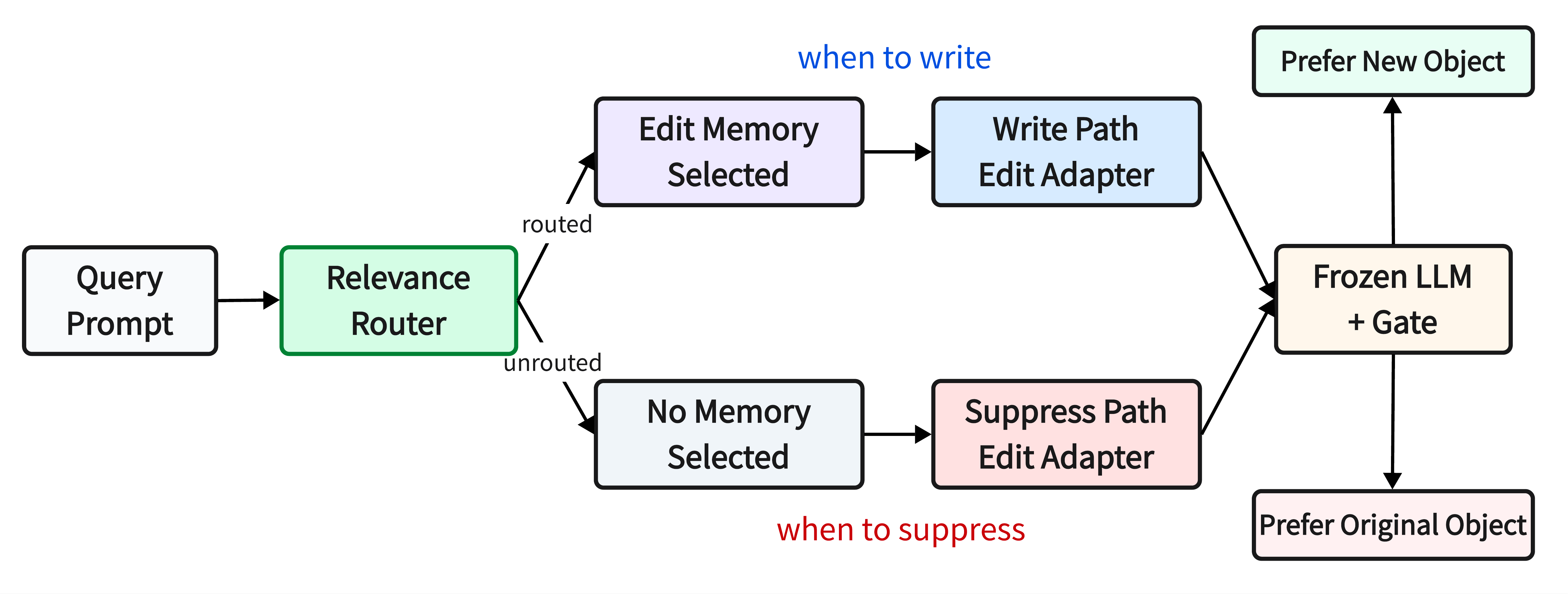}
\caption{Route-specialized memory-assisted editing. A relevance router decides whether to retrieve edit memory. Routed prompts use the edit adapter and are optimized to prefer the new object. Unrouted non-direct prompts use the locality adapter and are optimized to prefer the protected original object.}
\label{fig:pipeline}
\end{figure}

Figure~\ref{fig:pipeline} shows \method{}. The edit adapter handles direct prompts and routed non-direct prompts. It is trained with edited-answer cross-entropy and pairwise ranking:
\begin{equation}
  \mathcal{L}_{\mathrm{edit}}
  =
  \lambda_{\mathrm{ce}}\mathcal{L}_{\mathrm{ce}}
  +
  \lambda_{\mathrm{rank}}\mathcal{L}_{\mathrm{rank}},
\end{equation}
where
\begin{equation}
  \mathcal{L}_{\mathrm{rank}}
  =
  \softplus(m-\score(x,y^+)+\score(x,y^-)).
\end{equation}
For direct and paraphrase examples, $(y^+,y^-)=(\newobj,\oldobj)$. For locality examples used by the locality adapter, the preference is reversed.

The locality adapter is trained only on no-memory locality ranking examples:
\begin{equation}
  \mathcal{L}_{\mathrm{loc}}
  =
  \softplus(m_{\mathrm{loc}}-\score(x,\oldobj)+\score(x,\newobj)).
\end{equation}
At inference, the fixed route policy is:
\begin{equation}
(\phi,g)(x)=
\begin{cases}
(\phi_{\mathrm{edit}},0.75), & \text{direct prompt},\\
(\phi_{\mathrm{edit}},0.50), & \text{non-direct routed prompt},\\
(\phi_{\mathrm{loc}},0.75), & \text{non-direct unrouted prompt}.
\end{cases}
\end{equation}
The gates are selected on validation and then held fixed for all reported test results. This design makes the method deliberately simple: the module that writes an edit is not active when the router judges the prompt irrelevant.

\subsection{Baselines}

We compare with two resource-matched PEFT-style baselines under the same test items, memory bank, and router:
\begin{itemize}
  \item \textbf{LocFT-style}: a single LoRA adapter trained with direct edited-answer cross-entropy, inspired by the strong fine-tuning baseline in model editing \cite{yang2026locft}.
  \item \textbf{PRUNE-style}: a single LoRA adapter trained with direct cross-entropy plus locality KL regularization, inspired by perturbation-restrained editing \cite{ma2025prune}.
\end{itemize}
These are not official reproductions of LocFT-BF or PRUNE. They are local, resource-matched PEFT baselines designed to isolate the value of dual adapters under a shared memory protocol.

\section{Experiments}

\subsection{Setup}

Table~\ref{tab:protocols} summarizes the main protocols. All main runs use 1,000 edit cases per benchmark, 500 adapter steps, rank-8 LoRA, alpha 16, dropout 0.05, 4-bit model loading, and frozen base-model weights. The main Llama model is Llama-3.1-8B-Instruct; the robustness check uses Qwen3-8B. Experiments were run on a single 32GB GPU server. No paid external LLM API calls or annotation services were used.

\begin{table}[t]
\centering
\small
\caption{Main evaluation protocols. Counts are prompt counts before validation/test filtering. \mquake{} locality is synthetic cross-case locality in our probability-preference protocol, not the official \mquake{} exact-match generation evaluation.}
\label{tab:protocols}
\begin{tabular}{lcccl}
\toprule
Benchmark & Edit cases & Direct & Paraphrase & Locality / no-route \\
\midrule
\cf{} & 1,000 & 1,000 & 2,000 & 3,000 \cf{} neighborhoods \\
\zsre{} & 1,000 & 1,000 & 1,000 & 2,000 relation-specific locality prompts \\
\mquake{} & 1,000 & 1,000 & 2,000 & synthetic cross-case locality for editor eval \\
\bottomrule
\end{tabular}
\end{table}

For \cf{}, the test split contains 800 direct, 1,600 paraphrase, and 1,600 locality prompts, with no non-direct prompt overlap between training, validation, and test. \zsre{} and \mquake{} are converted to the same local schema and evaluated under the same memory protocol. We report direct, paraphrase, locality, overall accuracy, and the router's case-level correctness when applicable.

\subsection{Main Results Across Benchmarks and Base Models}

\begin{figure}[t]
\centering
\includegraphics[width=0.92\textwidth]{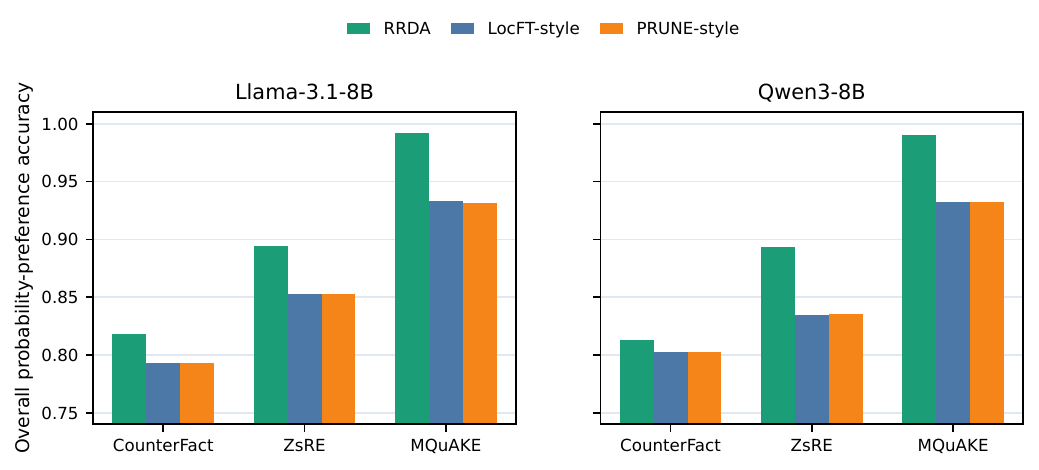}
\caption{Overall probability-preference accuracy on two base models. \method{} is best on all three benchmarks for both Llama-3.1-8B and Qwen3-8B.}
\label{fig:main-overall}
\end{figure}

Table~\ref{tab:main} gives the main results. \method{} has the highest overall accuracy on every benchmark for both base models. On Llama, \method{} improves over the better single-adapter baseline by 0.0250 on \cf{}, 0.0417 on \zsre{}, and 0.0594 on \mquake{}. On Qwen, the gains are 0.0105, 0.0583, and 0.0581. Figure~\ref{fig:main-overall} visualizes the same pattern.

\begin{table}[t]
\centering
\small
\caption{Main probability-preference results. All methods share the same split, memory bank, and selected router within each benchmark. Bold indicates the best overall result for each model-benchmark block.}
\label{tab:main}
\resizebox{\textwidth}{!}{
\begin{tabular}{ll l cccc}
\toprule
Base model & Benchmark & Method & Overall & Direct & Paraphrase & Locality \\
\midrule
\multirow{9}{*}{Llama-3.1-8B}
& \multirow{3}{*}{\cf{}} & \textbf{\method{}} & \textbf{0.8180} & 0.9975 & 0.6275 & 0.9188 \\
& & LocFT-style & 0.7928 & 0.9950 & 0.6063 & 0.8781 \\
& & PRUNE-style & 0.7930 & 0.9938 & 0.6088 & 0.8769 \\
\cmidrule{2-7}
& \multirow{3}{*}{\zsre{}} & \textbf{\method{}} & \textbf{0.8946} & 0.9675 & 0.8363 & 0.8800 \\
& & LocFT-style & 0.8529 & 0.9563 & 0.8400 & 0.7625 \\
& & PRUNE-style & 0.8525 & 0.9588 & 0.8363 & 0.7625 \\
\cmidrule{2-7}
& \multirow{3}{*}{\mquake{}} & \textbf{\method{}} & \textbf{0.9922} & 1.0000 & 0.9931 & 0.9825 \\
& & LocFT-style & 0.9328 & 0.9263 & 0.9225 & 0.9600 \\
& & PRUNE-style & 0.9319 & 0.9250 & 0.9213 & 0.9600 \\
\midrule
\multirow{9}{*}{Qwen3-8B}
& \multirow{3}{*}{\cf{}} & \textbf{\method{}} & \textbf{0.8133} & 0.9988 & 0.6381 & 0.8956 \\
& & LocFT-style & 0.8025 & 0.9950 & 0.6413 & 0.8675 \\
& & PRUNE-style & 0.8028 & 0.9975 & 0.6406 & 0.8675 \\
\cmidrule{2-7}
& \multirow{3}{*}{\zsre{}} & \textbf{\method{}} & \textbf{0.8933} & 0.9713 & 0.8375 & 0.8713 \\
& & LocFT-style & 0.8346 & 0.9663 & 0.8425 & 0.6950 \\
& & PRUNE-style & 0.8350 & 0.9663 & 0.8438 & 0.6950 \\
\cmidrule{2-7}
& \multirow{3}{*}{\mquake{}} & \textbf{\method{}} & \textbf{0.9906} & 1.0000 & 0.9913 & 0.9800 \\
& & LocFT-style & 0.9325 & 0.9288 & 0.9244 & 0.9525 \\
& & PRUNE-style & 0.9322 & 0.9288 & 0.9219 & 0.9563 \\
\bottomrule
\end{tabular}}
\end{table}

The gain is not uniform across benchmarks. On \cf{}, direct accuracy is already near saturated for all methods, and the main gain is locality: \method{} reaches 0.9188 on Llama versus about 0.878 for the baselines. On \zsre{}, the baselines have comparable paraphrase accuracy but much worse locality, which fits the same-subject different-relation failure mode. On \mquake{}, \method{} improves all three metrics, but this result should be interpreted within our composed-memory probability protocol.

\subsection{Memory Contribution Ablation}

Because \method{} combines explicit edit memory with route-specialized adapters, we further ask how much each source contributes. Table~\ref{tab:memory-ablation} compares four settings under the same selected router and test split. \textit{Base, no memory} evaluates the frozen base model without retrieved memory or adapters. \textit{Memory-only} uses the selected memory record with the frozen base model and no adapters. \textit{Adapter-only} uses the selected adapter route and gates but removes memory text from the prompt. \textit{Full \method{}} is the full method from Table~\ref{tab:main}.

\begin{table}[t]
\centering
\scriptsize
\caption{Memory contribution ablation. All rows use the benchmark-selected router. Overall is the micro-average over direct, paraphrase, and locality prompts.}
\label{tab:memory-ablation}
\resizebox{\textwidth}{!}{
\begin{tabular}{ll l cccc}
\toprule
Base model & Benchmark & Setting & Overall & Direct & Paraphrase & Locality \\
\midrule
\multirow{12}{*}{Llama-3.1-8B}
& \multirow{4}{*}{\cf{}} & Base, no memory & 0.4138 & 0.0838 & 0.1131 & 0.8794 \\
& & Adapter-only & 0.4280 & 0.1025 & 0.0988 & 0.9200 \\
& & Memory-only & 0.7858 & 0.9850 & 0.5938 & 0.8781 \\
& & Full \method{} & \textbf{0.8180} & 0.9975 & 0.6275 & 0.9188 \\
\cmidrule{2-7}
& \multirow{4}{*}{\zsre{}} & Base, no memory & 0.4375 & 0.2438 & 0.2488 & 0.8200 \\
& & Adapter-only & 0.4900 & 0.2738 & 0.2550 & 0.9413 \\
& & Memory-only & 0.8413 & 0.9413 & 0.8225 & 0.7600 \\
& & Full \method{} & \textbf{0.8946} & 0.9675 & 0.8363 & 0.8800 \\
\cmidrule{2-7}
& \multirow{4}{*}{\mquake{}} & Base, no memory & 0.2928 & 0.1675 & 0.1725 & 0.6588 \\
& & Adapter-only & 0.5278 & 0.5012 & 0.3825 & 0.8450 \\
& & Memory-only & 0.9234 & 0.9237 & 0.9050 & 0.9600 \\
& & Full \method{} & \textbf{0.9922} & 1.0000 & 0.9931 & 0.9825 \\
\midrule
\multirow{12}{*}{Qwen3-8B}
& \multirow{4}{*}{\cf{}} & Base, no memory & 0.4290 & 0.1038 & 0.1538 & 0.8669 \\
& & Adapter-only & 0.4340 & 0.1188 & 0.1300 & 0.8956 \\
& & Memory-only & 0.8020 & 0.9962 & 0.6394 & 0.8675 \\
& & Full \method{} & \textbf{0.8133} & 0.9988 & 0.6381 & 0.8956 \\
\cmidrule{2-7}
& \multirow{4}{*}{\zsre{}} & Base, no memory & 0.4438 & 0.2913 & 0.2975 & 0.7425 \\
& & Adapter-only & 0.5142 & 0.3088 & 0.3063 & 0.9275 \\
& & Memory-only & 0.8283 & 0.9600 & 0.8313 & 0.6938 \\
& & Full \method{} & \textbf{0.8933} & 0.9713 & 0.8375 & 0.8713 \\
\cmidrule{2-7}
& \multirow{4}{*}{\mquake{}} & Base, no memory & 0.3122 & 0.1913 & 0.1981 & 0.6613 \\
& & Adapter-only & 0.5112 & 0.4725 & 0.3694 & 0.8337 \\
& & Memory-only & 0.9272 & 0.9213 & 0.9206 & 0.9463 \\
& & Full \method{} & \textbf{0.9906} & 1.0000 & 0.9912 & 0.9800 \\
\bottomrule
\end{tabular}}
\end{table}

The ablation shows that explicit memory is the dominant source of edit reliability and paraphrase generalization: memory-only already raises overall accuracy from 0.4138 to 0.7858 on Llama \cf{}, from 0.4375 to 0.8413 on Llama \zsre{}, and from 0.2928 to 0.9234 on Llama \mquake{}. The same pattern holds for Qwen. However, memory alone is not the final solution. On \zsre{}, memory-only hurts locality because same-subject relation-specific locality prompts can still be exposed to an irrelevant edit; the full policy recovers locality from 0.7600 to 0.8800 on Llama and from 0.6938 to 0.8712 on Qwen. Adapter-only results further show that the route-conditioned adapter path is not storing all edited facts by itself, but it does improve locality and, on \mquake{}, contributes nontrivially to direct and paraphrase preferences. Thus the main gain comes from explicit memory, while the dual-adapter route policy improves the reliability-locality balance needed for the full result.

\subsection{Router Ablation}

\begin{figure}[t]
\centering
\includegraphics[width=0.95\textwidth]{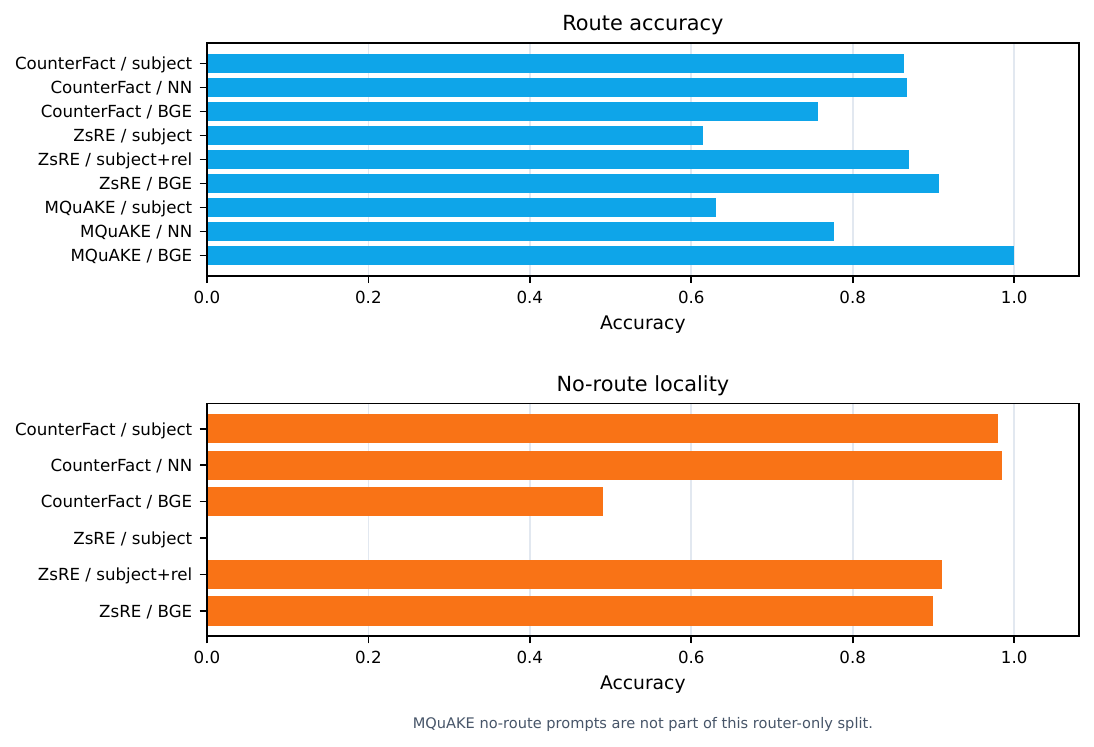}
\caption{Different relevance boundaries are needed across benchmarks. \cf{} selects NN because it preserves no-route locality, while \zsre{} and \mquake{} select BGE semantic routing.}
\label{fig:router}
\end{figure}

Table~\ref{tab:router} and Figure~\ref{fig:router} show why router choice is part of the method rather than a replaceable detail. The phrase ``different relevance boundaries'' means that the same surface cue is not safe for every dataset: \cf{} rewards routers that avoid over-routing locality prompts, \zsre{} requires relation-sensitive routing because locality prompts may share the subject, and \mquake{} needs semantic retrieval for multi-hop questions. On \cf{}, BGE has the highest pair F1, but it over-routes neighborhoods and drops no-route locality to 0.4900; we therefore select NN because it has the best route accuracy and no-route locality. On \zsre{} and \mquake{}, BGE is selected because it gives the best route accuracy.

\begin{table}[t]
\centering
\small
\caption{Router ablation. The selected router is used in the main editor evaluation. Dashes indicate that a no-route locality split is not part of the router-only \mquake{} protocol.}
\label{tab:router}
\begin{tabular}{llccccc}
\toprule
Benchmark & Router & Selected & Pair F1 & Route acc. & Direct & No-route \\
\midrule
\cf{} & subject & no & 0.8129 & 0.8633 & 0.9900 & 0.9800 \\
\cf{} & NN & yes & 0.8258 & 0.8667 & 1.0000 & 0.9850 \\
\cf{} & BGE & no & 0.8889 & 0.7567 & 1.0000 & 0.4900 \\
\midrule
\zsre{} & subject & no & 0.6143 & 0.6150 & 0.9550 & 0.0000 \\
\zsre{} & subject+relation & no & 0.8632 & 0.8700 & 0.9750 & 0.9100 \\
\zsre{} & BGE & yes & 0.8849 & 0.9067 & 1.0000 & 0.9000 \\
\midrule
\mquake{} & subject & no & 0.7740 & 0.6300 & 0.6200 & -- \\
\mquake{} & NN & no & 0.8590 & 0.7767 & 0.8133 & -- \\
\mquake{} & BGE & yes & 0.9926 & 1.0000 & 1.0000 & -- \\
\bottomrule
\end{tabular}
\end{table}

We also ran a \cf{} stress set with lexical distractors, same-subject different-relation negatives, and multi-subject prompts. Subject-only routing reaches only 0.7925 route accuracy because it routes distractors whenever the subject appears. The NN router reaches about 0.9900 route accuracy on this stress set. This motivates hybrid future routers: semantic retrieval for recall plus supervised calibration for no-route boundaries.

\subsection{Component Ablation}

Table~\ref{tab:components} isolates the two adapters. On \cf{}, editor-only has stronger paraphrase behavior than suppressor-only but worse locality; suppressor-only has better locality but worse direct and paraphrase. The full dual policy combines the relevant strengths. \zsre{} is even clearer: the suppressor-only row has high locality but lower edit accuracy, while the dual policy recovers a better overall balance. \mquake{} is the outlier because the current semantic router routes almost all evaluated prompts into the edit path; therefore the locality adapter is rarely used.

\begin{table}[t]
\centering
\small
\caption{Component ablation on Llama-3.1-8B. The dual row is the corresponding main-result \method{} run.}
\label{tab:components}
\begin{tabular}{llcccc}
\toprule
Benchmark & Component & Overall & Direct & Paraphrase & Locality \\
\midrule
\cf{} & editor only & 0.8038 & 0.9975 & 0.6325 & 0.8781 \\
\cf{} & suppressor only & 0.7998 & 0.9850 & 0.5888 & 0.9181 \\
\cf{} & dual policy & \textbf{0.8180} & \textbf{0.9975} & 0.6275 & \textbf{0.9188} \\
\midrule
\zsre{} & editor only & 0.8588 & \textbf{0.9675} & \textbf{0.8463} & 0.7625 \\
\zsre{} & suppressor only & 0.8788 & 0.9413 & 0.8138 & \textbf{0.8813} \\
\zsre{} & dual policy & \textbf{0.8946} & \textbf{0.9675} & 0.8363 & 0.8800 \\
\midrule
\mquake{} & editor only & \textbf{0.9922} & \textbf{1.0000} & \textbf{0.9931} & \textbf{0.9825} \\
\mquake{} & suppressor only & 0.9238 & 0.9238 & 0.9050 & 0.9613 \\
\mquake{} & dual policy & \textbf{0.9922} & \textbf{1.0000} & \textbf{0.9931} & \textbf{0.9825} \\
\bottomrule
\end{tabular}
\end{table}

\begin{figure}[t]
\centering
\includegraphics[width=0.55\textwidth]{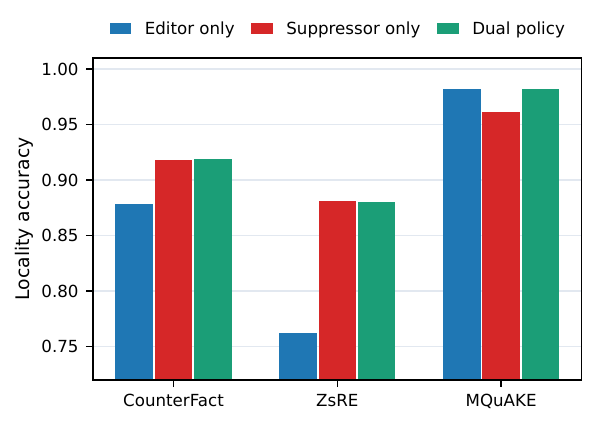}
\caption{Locality in the component ablation. The locality adapter is most informative on \cf{} and \zsre{}, where off-route prompts are present and nontrivial.}
\label{fig:components}
\end{figure}

\subsection{Target Module Ablation}

\begin{figure}[t]
\centering
\includegraphics[width=0.56\textwidth]{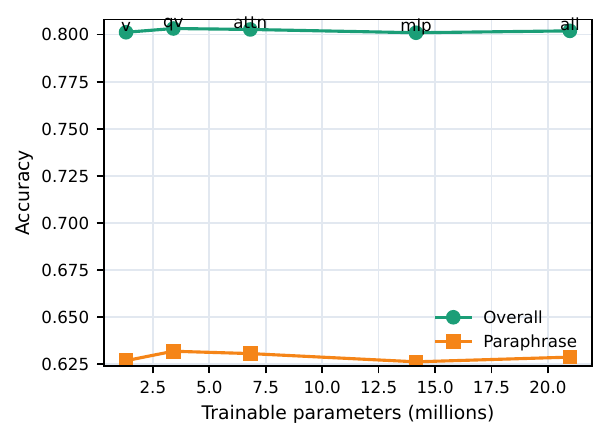}
\caption{More target modules do not materially improve \cf{} accuracy under the rank-8, 500-step setting.}
\label{fig:module}
\end{figure}

We next test whether the method's gain comes simply from more LoRA target modules. Table~\ref{tab:module} reports a rank-8, 500-step \cf{} target-module sweep. We intentionally report the practically relevant target sets used in the main design. q/v reaches the best overall among the shown settings, while attention-all, MLP, and all-module LoRA add many more trainable parameters without improving accuracy. Figure~\ref{fig:module} shows the same pattern.

\begin{table}[t]
\centering
\small
\caption{Rank-8, 500-step target-module ablation on \cf{}. The main method uses q/v as a compact default.}
\label{tab:module}
\begin{tabular}{lccccc}
\toprule
Target set & Trainable params & Overall & Direct & Paraphrase & Locality \\
\midrule
v & 1.31M & 0.8013 & 0.9975 & 0.6269 & 0.8775 \\
q/v & 3.41M & \textbf{0.8033} & 0.9975 & \textbf{0.6319} & 0.8775 \\
attention & 6.82M & 0.8028 & 0.9975 & 0.6306 & 0.8775 \\
MLP & 14.16M & 0.8010 & 0.9975 & 0.6263 & 0.8775 \\
all modules & 20.97M & 0.8020 & 0.9975 & 0.6288 & 0.8775 \\
\bottomrule
\end{tabular}
\end{table}

\section{Discussion}

The experiments support three lessons. First, the boundary for relevant memory is dataset dependent. A subject router can be strong on clean \cf{}, but it fails relation-specific locality on \zsre{} and lexical distractors in stress tests. A semantic router improves \zsre{} and \mquake{}, but can over-route \cf{} neighborhoods. Second, explicit memory is necessary but insufficient. The memory-only ablation explains most of the reliability and paraphrase gain, yet it can damage locality on relation-specific prompts; the full dual policy restores that balance. Third, edit injection and locality preservation are different operations. The edit adapter is useful when the memory is relevant; the locality adapter is useful when the system should avoid using the edit. Simply increasing LoRA capacity is not enough. The strongest evidence is the memory and component ablations, where memory, edit adaptation, and locality suppression have visibly different roles.

The method should therefore be read as a route-specialized design pattern rather than a claim of universal state of the art. It is attractive because it is deterministic, cheap, and auditable: memory records are explicit, the router can be inspected, and the active adapter is known for each prompt. This is useful in applications where edit provenance matters and where silent probability leakage is unacceptable.

\section{Limitations}

There are several important limitations. First, the baselines are resource-matched PEFT-style implementations, not official reproductions of LocFT-BF, PRUNE, ROME, MEMIT, MEND, SERAC, or MeLLo under EasyEdit. Second, \mquake{} is evaluated with composed edit memory and probability preference plus synthetic locality; it is not the official \mquake{} exact-match generation benchmark. Third, \cf{} paraphrase accuracy remains low, around 0.63, even though direct and locality accuracy are high. This suggests that paraphrase augmentation or stronger semantic routing is still needed. Fourth, the current semantic router is not calibrated enough for \cf{} no-route locality. Fifth, all main results use 7B/8B instruction models and short 500-step adapter training; larger models and longer sequential-edit settings may reveal different failure modes. Finally, the method can edit model behavior in ways that may be socially sensitive, so deployment should require edit authorization, provenance logging, and post-edit audits.

\section{Conclusion}

We presented \method{}, a route-specialized dual-adapter editor for memory-assisted knowledge editing. The method asks two separate questions: when should an edit be written, and when should its effect be suppressed? A relevance router chooses memory; an edit adapter handles routed prompts; a locality adapter handles unrouted non-direct prompts. Across \cf{}, \zsre{}, and \mquake{}, and across Llama-3.1-8B and Qwen3-8B, this simple separation gives the best overall probability-preference accuracy among the tested method families. The ablations show that explicit memory provides the largest edit signal, but the full gain requires separating edit injection from off-route locality suppression rather than merely increasing LoRA capacity.

\bibliographystyle{splncs04}
\bibliography{references}

\end{document}